\documentclass[11pt,twocolumn,letterpaper]{article}
\usepackage[pagenumbers]{wacv} 

\usepackage{amsmath}
\usepackage{amssymb}
\usepackage{booktabs}
\usepackage{adjustbox}
\usepackage[hidelinks]{hyperref}
\usepackage{balance}
\usepackage[table]{xcolor}

\begin{document}

\title{Predicting Space Tourism Demand Using Explainable AI}

\author{Tan-Hanh Pham, Jingchen Bi, Rodrigo Mesa-Arango, and Kim-Doang Nguyen$^*$\\
\hphantom{text}\\
Florida Institute of Technology, USA\\
\hphantom{text}\\
}

\maketitle

\begin{abstract}
Comprehensive forecasts of space tourism demand are crucial for businesses to optimize strategies and customer experiences in this burgeoning industry. Traditional methods struggle to capture the complex factors influencing an individual's decision to travel to space. In this paper, we propose an explainable and trustworthy artificial intelligence (AI) framework to address the challenge of predicting space tourism demand by following the National Institute of Standards and Technology guidelines. We develop a novel machine learning network, called SpaceNet, capable of learning wide-range dependencies in data and allowing us to analyze the relationships between various factors such as age, income, and risk tolerance. In particular, the space travel demand for people living in the US is investigated, in which we frame the demand into four travel types: no travel, moon travel, suborbital, and orbital travel. To this end, we collected 1860 data points in many states and cities with different ages and then conducted our experiment with the data. 

\def\thefootnote{$\$$}\footnotetext{This material is based upon work supported by the U.S. National Science Foundation under Grant \#2245022. }

As a result, our SpaceNet model achieves an average Receiver Operating Characteristic Area Under the Curve or ROC-AUC of 0.82 $\pm$ 0.088, which indicates a good performance for the model's classification. Our investigation demonstrated that travel price, age, annual income, gender, and fatality probability are important features in deciding whether a person wants to travel or not. Beyond demand forecasting, we use explainable AI to provide interpretation for the travel-type decisions of an individual, offering insights into the factors driving interest in space travel, which is not possible with traditional classification methods. This knowledge enables businesses to tailor marketing strategies and optimize service offerings in this rapidly evolving market. To the best of our knowledge, this is the first work to implement an explainable and interpretable AI framework for investigating the factors influencing space tourism.
\end{abstract}

\section{Introduction}
\label{sec.1intro}

\subsection{Background}
\label{literature-review}
The dream of flight has captivated humanity for centuries, inspiring visionaries to turn imagination into reality. One of the earliest dreamers, Leonardo da Vinci, sketched flying machines that paved the way for the Wright brothers' historic first powered flight in 1903 \cite{crouch2003bishop}. This revolutionary achievement marked the dawn of the aviation era, forever transforming transportation and igniting a passion for reaching new heights.  Commercially viable airplanes soon followed, as evidenced by the first paying passenger flight in 1914 by Tony Jannus, who piloted a Benoist Type XIV flying boat from St. Petersburg to Tampa, Florida \cite{nasa_airplanes}. Air travel rapidly evolved, fostering global connections and propelling tourism to unprecedented levels. 

Human ambition, however, did not stop at the atmosphere's edge. The desire to explore the cosmos drove advancements in space exploration technologies. Pioneering efforts by countries including the United States and the Soviet Union during the Space Race culminated in Yuri Gagarin’s groundbreaking orbital flight in 1961, making him the first human in space \cite{burgess2022soviets, neufeld2018spaceflight}. Shortly after, the United States achieved its own milestone with John Glenn’s successful orbital flight aboard Friendship 7 in 1962, proving American capabilities in human spaceflight \cite{burgess2015friendship}. This era marked remarkable advancements in rocketry and spacecraft design, establishing the foundation for future space travel.

In recent years, the landscape of space exploration has further evolved with the emergence of private space companies drastically changing the game for space travel. This shift is exemplified by NASA's collaboration with private companies like SpaceX under the Commercial Orbital Transportation Services (COTS) program, established in 2006 \cite{mazzucato2018co}. The COTS aimed to develop private capabilities for resupplying the International Space Station (ISS). This partnership not only marked a significant step towards commercial spaceflight but also laid the groundwork for private companies to become major players in the flourishing space tourism industry \cite{lindenmoyer2015commercial}. Building on this foundation, space travel has transitioned from the realm of science fiction to reality. In particular, reusable launch vehicles from companies like SpaceX have revolutionized the space transportation landscape, enabling more affordable and frequent missions. For example, on October 14, 2024, SpaceX succeeded in bringing part of its giant Starship rocket back to its launch pad, this shift has opened doors for private space companies to thrive. In addition to SpaceX, Blue Origin, Virgin Galactic, Sierra Space, World View, and Axiom Space are notable companies at the forefront of developing innovative solutions for space tourism experiences.

Examining the current landscape of space tourism, two critical questions arise: who are the potential space tourists, and what drives their decisions? Understanding their motivations and preferences is key to shaping the future of this burgeoning industry. While historically, space travel was an exclusive privilege reserved for government-funded astronauts, a recent shift is underway. Starlust highlights a rise of space companies and a significant decrease in spaceflight costs \cite{Start}. This trend towards affordability is reflected in the pricing structures offered by these new players. SpaceX, for instance, caters to both suborbital and orbital experiences, providing a range of space tourism options. While suborbital flights offer a few minutes of weightlessness and breathtaking views of Earth, orbital flights extend the experience, potentially involving multi-day stays in space stations. Currently, suborbital flights are the primary focus, but the exploration of orbital space tourism is actively underway. In 2022, SpaxeX provides a 10-day orbital trip to ISS at the price of \$55 million \cite{Start}. Virgin Galactic is another major player, offering suborbital flights for a price tag of \$450,000 \cite{blueorigin}. While Blue Origin also offers suborbital experiences, their pricing is estimated between \$200,000 and \$300,000.

In July 2021, Virgin Galactic took a significant step towards commercial spaceflight with its founder, Richard Branson, piloting a test flight to the edge of Earth's atmosphere alongside two pilots and three other company employees. Shortly thereafter, in the same month, Blue Origin's New Shepard rocket carried Jeff Bezos, Mark Bezos, Oliver Daemen (a young auction winner who paid approximately \$28 million for the experience), and Wally Funk (a pioneering female aviator from the Mercury 13 program) on a suborbital spaceflight. These successful missions mark a pivotal moment in the growing space tourism industry.

\subsection{Literature Review}

The research interest in space travel has been amplified since the early days of the Space Age in the 1950s. It has explorationally expanded after the successful human spaceflight missions by commercial companies, for example, SpaceX's groundbreaking launch in 2020, which sent NASA astronauts to the International Space Station. This milestone, along with Virgin Galactic's and Blue Origin's subsequent suborbital flights in 2021, has significantly broadened the scope of space tourism research. Scholars have explored various dimensions of this new frontier, focusing on understanding the factors that drive consumer behavior in space tourism. In particular, work in \cite{laing2004australian} identified the cost, safety, and product design, along with various demographic and behavioral characteristics, significantly influence individuals' attitudes and interests in space tourism. Building on this, their follow-up study \cite{laing2005extraordinary} further indicated that factors such as attitudes toward challenges, goal-setting patterns, need for self-actualization, customers' novelty-seeking and adventurousness, and childhood experiences are also linked to motivations for engaging in space tourism. In 2009, \cite{crouch2009modelling} conducted discrete choice experiments to examine consumer preferences across four categories of space tourism: high-altitude jet fighter flights, atmospheric zero-gravity flights, short-duration sub-orbital flights, and longer-duration orbital trips. Their findings reinforced the notion from prior studies that prospective space tourists are highly price-sensitive. Additionally, they confirmed that the type of space flights and specific features of space tourism products significantly influence customer choices.

\begin{figure*}[h]
    \centering
    \small
    \begin{subfigure}[b]{0.45\textwidth}
        \centering
        \includegraphics[width=\textwidth]{ 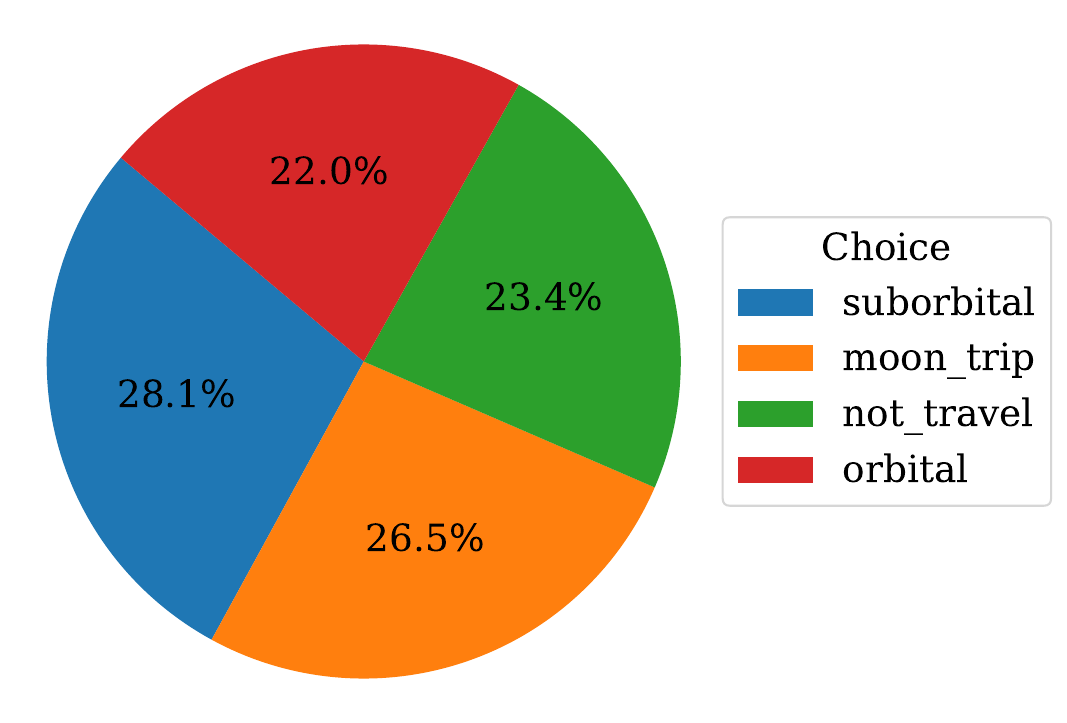}
        \caption{Space travel options.}
        \label{fig.spacetravel}
    \end{subfigure}
    \begin{subfigure}[b]{0.45\textwidth}
        \centering
        \includegraphics[width=\textwidth]{ 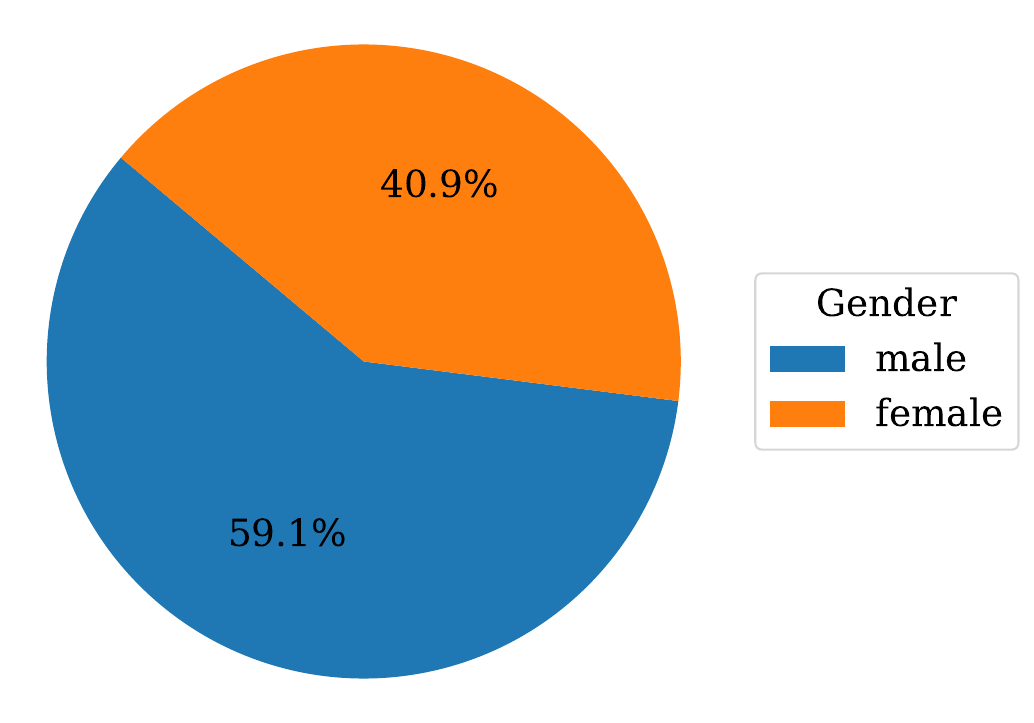}
        \caption{Gender distribution.}
        \label{fig.gender}
    \end{subfigure}
    \vspace{1em} 
    \begin{subfigure}[b]{0.45\textwidth}
        \centering
        \includegraphics[width=\textwidth]{ 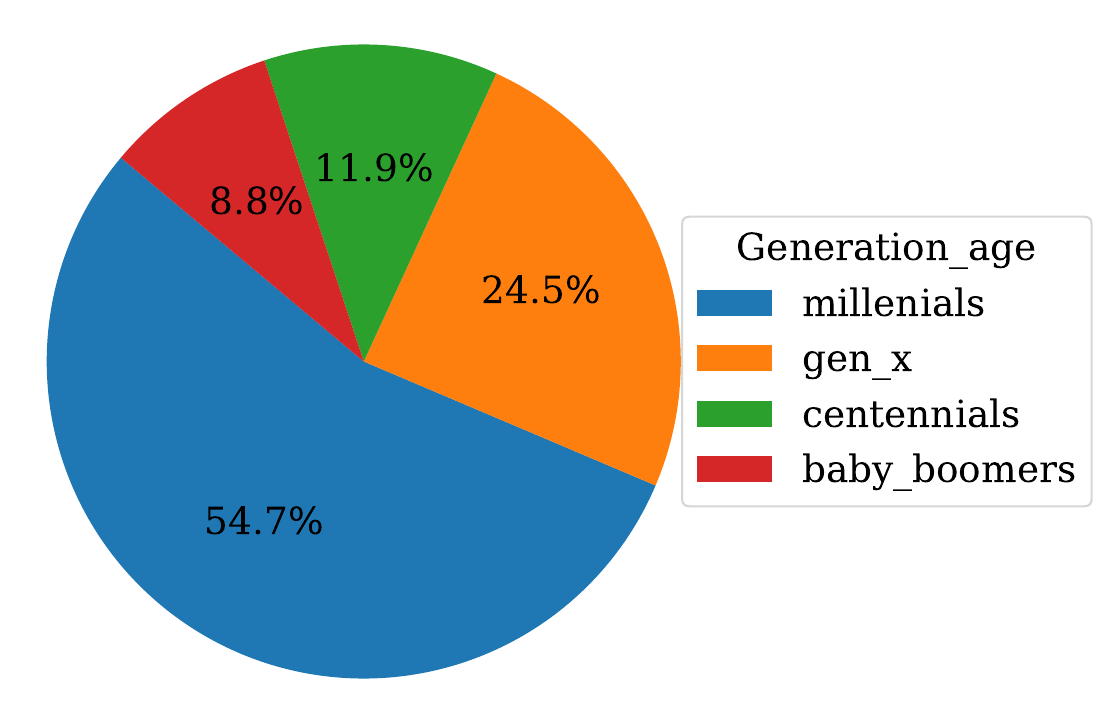}
        \caption{Generation distribution.}
        \label{fig.Generationage}
    \end{subfigure}
    \begin{subfigure}{0.45\textwidth}
        \centering
        \includegraphics[width=\textwidth]{ 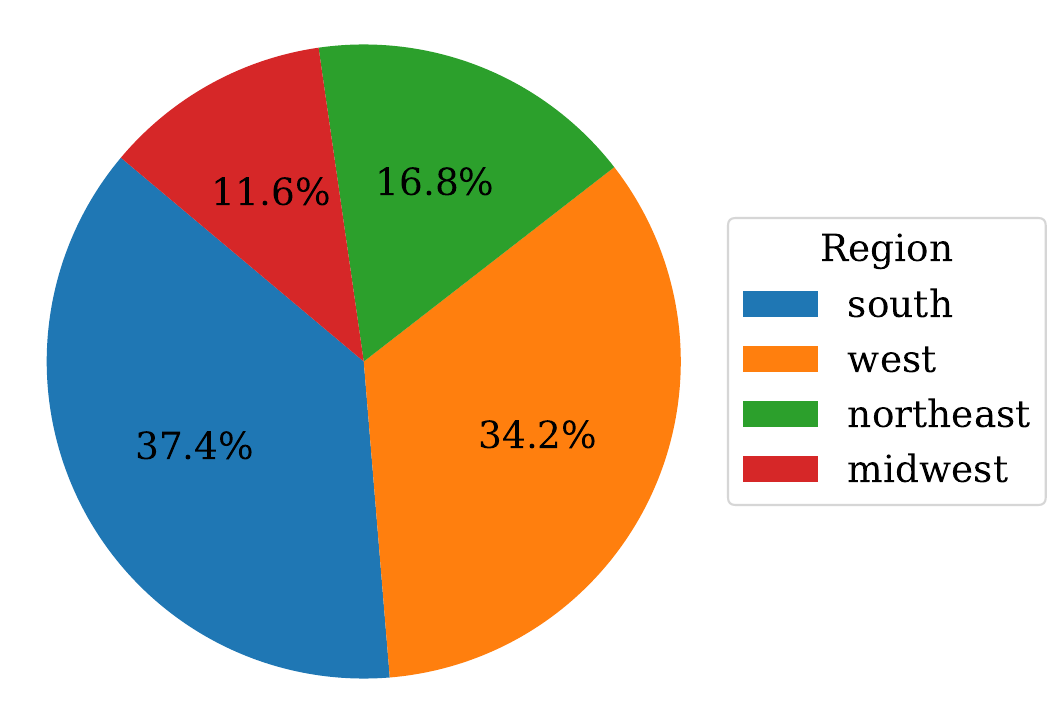}
        \caption{Regional distribution in the U.S.}
        \label{fig.region}
    \end{subfigure}

    \caption{Distribution of representative attributes in our collected data.}
    \label{fig.DataRepresentation}
\end{figure*}

The study by \cite{reddy2012space} aims to examine the perceptions of potential space travelers regarding key factors that influence their motivation, behavior, and decision-making. The methodology includes quantitative data collection and qualitative interviews with key informants from the space tourism industry. The analysis reveals that the intrinsic need for adventure and exploration is a primary motivational force in space tourism, with perceived risk significantly influencing the willingness to undertake space travel. In 2023, work in \cite{le2013astrium} conducted a comprehensive market analysis supported by two global surveys to evaluate the viability of the space tourism market. Using a logic model with a binary choice, they calculated the conditional probability of taking a suborbital flight. The findings indicate that the primary motivation for spending on space tourism is the excitement of participating in a rare adventure, rather than for sightseeing or special experiences. While environmental and safety concerns may deter some Europeans, they have less impact on potential travelers from the United States or China. Similarly, several studies further investigating the impact of attitudes, risk factors, perceived risk, financial cost, and desired behavioral intentions of space travelers were studied \cite{chang2017preliminary, olya2020antecedents, olya2023emerging, wang2022impact}.

With the development of machine learning, its application has made significant contributions in areas such as data analysis, image recognition, and natural language processing, thus facilitating various applications across diverse domains, including healthcare, finance, and marketing \cite{lecun2015deep, shrestha2019review, vaswani2017attention, touvron2023llama, team2023gemini, pham2025enhanced, pham2024soil, pham2024adaptive, achiam2023gpt, pham2024deep, pham2023seunet}. In the space tourism domain, \cite{zyma2022application} analyzed appealing development directions for tourism companies using machine learning models and socioeconomic data. The study identified Italy, Spain, Croatia, Greece, Portugal, and Poland as the most promising markets for tourism businesses. These countries offer high safety, moderate costs, low market entry barriers, steady industry growth, favorable business environments, attractive natural resources, and well-developed infrastructure.

Subsequently, \cite{mesa2023missions} investigated factors influencing travelers' choices using data collected in the US and machine learning models. Their findings revealed that risk avoidance is the primary deterrent for space tourism, particularly among older individuals, women, and households with children. Moreover, \cite{gatti2023assessing} applied linear regression analysis to explore the intentions and traits of future space travelers. Their findings showed that scores on the Space Tourism Propensity Questionnaire (STP-Q) are influenced by personality traits, including sensation seeking, social distance, and conscientiousness. Furthermore, \cite{kim2023space} investigated perceived risks in space tourism, highlighting age and gender differences using a fuzzy-set Qualitative Comparative Analysis. Their findings revealed significant variations in perceived risks between older and younger individuals, as well as between males and females. In a subsequent study, they examined factors influencing participation in orbital and suborbital space tourism, employing various analytical methods, including deep learning \cite{kim2024orbital}. The results indicated that extrinsic motivation and trust in artificial intelligence strongly influence behavioral intention, while intrapersonal constraints negatively impact participation. Interpersonal constraints were found to be insignificant in some analyses but notable in others.

\subsection{Research Gaps and Motivations}
While some research has explored consumer behavior in space tourism, gaps remain. There is a lack of comprehensive studies integrating various risk factors, including psychological, financial, and safety-related, and their combined impact on consumer behavior. Most existing research provides only a snapshot of preferences or intentions at a specific time, missing longitudinal studies that track changes in attitudes over time, especially in the future.  Additionally, although regional differences are sometimes considered, there is limited research on space tourism demand in the United States. These limitations hinder their application in safety-critical domains, as outlined by the NIST guidelines for trustworthy AI \cite{phillips2021four}. Furthermore, given the high costs associated with space travel, understanding pricing structures is vital for the growth and sustainability of the space tourism industry. Prices, currently ranging from several hundred thousand to millions of dollars, significantly influence the decision-making process of potential space tourists.

To address the problem, this paper investigates the importance of comprehensively analyzing the factors to better forecast demand and tailor offerings to meet market needs. This is achieved by developing a novel explainable artificial intelligence (xAI) framework to extract meanings out of quantitative human subject data. By focusing on explainable and interpretable techniques, the work provides insights into the inner workings of the machine learning model and the reasoning behind decisions more transparently and understandably.

Previous studies show that xAI has the potential to significantly impact various scientific endeavors. It allows researchers to assess the validity of model predictions and uncover new scientific insights in the medical prediction \cite{knapivc2021explainable}. For example, recent work by \cite{nordin2023explainable} demonstrates how it can be used to explain risk factors in predicting suicide attempts. Similarly, \cite{sariyer2024leveraging} explored the techniques for public transportation and tourism prediction, highlighting its potential for understanding user decision-making in choosing between ferries, railways, and buses. Additionally, \cite{davazdahemami2022explanatory} investigated its effectiveness in identifying and validating risk factors of novel diseases, while \cite{nimmy2022explainability} applied the technique to determine operational risks, aiding supply chain risk managers in planning and development.

\subsection{Novel Contributions}

Inspired by these research gaps and remained challenges of the space tourism, our study develops a trustworthy and explainable AI framework for space tourism human subject data extraction, offering valuable insights into user preferences and decision-making in this emerging field. Our original contributions are summarized as follows:

\begin{itemize}
    \item We propose a novel deep learning model, SpaceNet, for extracting meanings from space travel data.
    \item We propose a novel explainable AI framework tailored for space tourism demand prediction, ensuring transparency and reliability in AI-driven decision-making processes.
    \item Our framework accurately predicts space tourism demand across various travel types, demonstrating robustness across different demographic groups.
    \item We provide a detailed analysis of the key factors influencing space tourism decisions, offering valuable insights for businesses to refine marketing strategies and optimize service offerings.
    \item To the best of our knowledge, this is the first study to implement an explainable and interpretable AI framework for investigating the factors influencing space tourism.
\end{itemize}

\section{Dataset}
\label{sec.data}

One crucial aspect of trustworthy AI is ensuring fairness and mitigating the potential for harmful biases. Traditional "black box" AI models can perpetuate biases present in the data they are trained on, leading to discriminatory outcomes where the model unfairly disadvantages certain groups. For example, in the context of space tourism, an AI model predicting demand might unintentionally favor potential travelers from wealthier regions due to historical data biases. This could result in overlooking enthusiastic participants from less affluent areas. Therefore, we conducted experiments to mitigate bias in data collection and preprocessing, by using strategies as follows:
\begin{itemize}
    \item Ensuring the diversification and representation of training data: Utilize datasets from various sources to capture a broader representation of the population that the model will be applied to. 
    \item Data cleaning and filtering: Identify and remove data points containing biases (e.g., removing features if not relevant to the model's purpose). Identify the most important features for building an xAI model. 
    \item Data augmentation: Create synthetic data to fill in gaps or underrepresented groups in the original dataset. 
\end{itemize}

\begin{figure*}[h]
    \centering
    \includegraphics[width=0.9\linewidth]{ 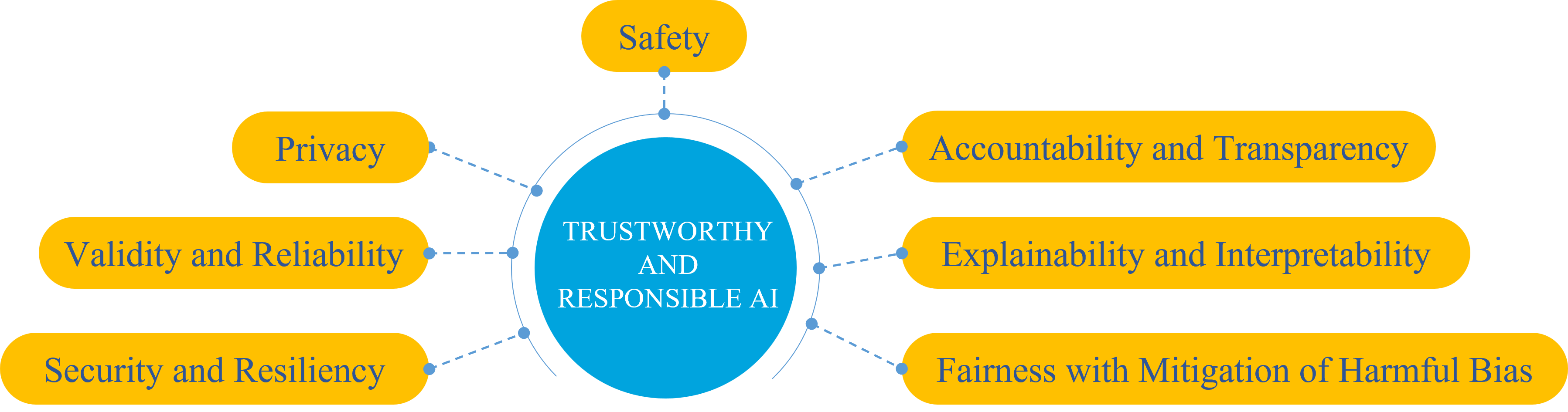}
    \caption{Trustworthy and responsible AI.}
    \label{TrustworthyAI}
\end{figure*}

\subsection{Data Diversification and Representation}
We collected data and ensured diversity by incorporating forty-four different attributes, including age, region, annual income, travel price, and level of education, among others. Figure \ref{fig.DataRepresentation} presents some representative attributes from our survey on interest in space tourism, conducted among a diverse population in the United States. The survey asked participants to choose their preferred travel option from four choices: orbital travel, suborbital travel, travel to the moon, and no interest in space travel, as illustrated in Fig. \ref{fig.DataRepresentation}. The number of data points used in our study is 1860 samples.

Following our survey, the distribution of interest across the four travel types indicates a relatively balanced preference. In Fig. \ref{fig.spacetravel}, suborbital travel refers to a trajectory where a spacecraft reaches space but does not achieve the necessary velocity to remain in orbit around the Earth. Instead, it follows a curved path, reaching the edge of the atmosphere and then descending back to the surface. Orbital travel, on the other hand, involves a spacecraft reaching a velocity high enough to enter a stable orbit around the Earth, typically exceeding 17,500 miles per hour (28,000 kilometers per hour). This allows the spacecraft to continuously circle the planet without falling back to the surface, enabling longer missions. 

Among those interested, suborbital travel appears to be the most popular choice nationally, with approximately 28.1\% of total responses followed by moon travel with an interest of 26.5\%. Orbital travel obtained an interest of approximately 22\% of respondents, while people who do not want to travel occupy a portion of approximately 23.4\%.

The data also incorporates demographic information, including participant gender (male and female), and age generation ranges from ``Baby Boomers" to ``Millennials" as illustrated in Fig.~\ref{fig.gender} and Fig.~\ref{fig.Generationage}, respectively. Among these groups, Millennials and males occupy the largest portion of the population in our survey. Furthermore, to understand the space travel demand, the survey is conducted across different regions in the US (West, South, Midwest, and Northeast) with geographic locations across over 30 states in the US. 


\subsection{Data Cleaning and Filtering}
\label{subsec.data_cleaning}

Our commitment to ensuring fairness in the AI model predicting demand for space tourism extends beyond diversifying data sources.  We acknowledge the importance of data cleaning and filtering to mitigate potential biases present within the collected data. Additionally, data augmentation techniques can be employed to address the under-representation of specific groups.

One crucial step in mitigating bias is identifying and removing outliers or data points that skew the results.  These outliers can arise from various sources, such as human error during data entry or inherent limitations of data collection methods. Therefore, we propose utilizing the Interquartile Range (IQR) technique for outlier detection \cite{dekking2005modern}. IQR represents the middle 50\% of the data distribution. It is calculated as the difference between the third quartile ($Q3$) and the first quartile ($Q1$): 
\begin{equation}
\text{IQR} = Q3 - Q1.
\label{eq:iqr}
\end{equation}
Data points falling outside a specific range, typically $1.5 \times \text{IQR}$ below the first quartile or above the third quartile, can be considered outliers and warrant further investigation. Specifically, a data point $x$ is an outlier if: 
\begin{equation}
x < Q1 - 1.5 \times \text{IQR} \quad \text{or} \quad x > Q3 + 1.5 \times \text{IQR}.
\label{eq:outlier}
\end{equation}

For instance, an outlier in the average travel price data might be a respondent reporting a significantly higher or lower travel cost compared to the rest of the participants. Investigating such outliers can help us determine if they represent data points, perhaps reflecting a luxury space-tourism package or a data entry error. By removing outliers that deviate significantly from the IQR range, we ensure the model is trained on a more accurate representation of average travel prices for space tourism. This helps minimize the influence of extreme values on the model's predictions. Following this process, we apply the standard normalization for all attributes in the dataset.

Beyond data cleaning, identifying the most relevant features is crucial for building an accurate and interpretable AI model for assessing space-tourism demand. In this study, we leverage the strengths of the Random Forest Classifier algorithm to achieve this objective \cite{breiman2001random}. Random Forest Classifiers operate by constructing an ensemble of decision trees. Each tree makes a prediction based on a subset of randomly selected characteristics. The final prediction is determined by aggregating the votes from all the individual trees in the forest. This ensemble approach offers several advantages, including robustness to overfitting and the ability to handle high-dimensional data, which can be the case when dealing with a variety of factors influencing space tourism demand.

Mathematically, let $\mathcal{D}$ be the training dataset and $\theta$ be a random vector used for selecting a subset of the features and samples. For each tree $T$ in the forest, the training dataset $\mathcal{D}$ is bootstrapped to create $\mathcal{D}_\theta$. Each tree is built using $\mathcal{D}_\theta$, and the prediction for a sample $x$ is given by $h(x, \theta)$, where $h(x, \theta)$ is the prediction of tree $T$ for input $x$. The final prediction of the Random Forest is obtained by averaging the predictions of majority voting:

\begin{equation}
    \label{}
    H(x) = \frac{1}{N} \sum_{i=1}^{N} h(x, \theta_i)
\end{equation}

The key strength of Random Forest Classifiers lies in their ability to estimate the relative importance of each feature in making predictions. This feature importance analysis is based on the concept of impurity, which refers to the randomness or heterogeneity within a set of data points. For classification tasks, the Gini impurity metric is commonly used. The Gini impurity for a node $t$ is calculated as:

\begin{equation}
    \label{}
    G(t) = 1 - \sum_{i=1}^{C} p_i^2,
\end{equation}
where $C$ is the number of classes and $p_i$ is the proportion of samples belonging to class $i$ in node $t$. When a feature $f$ is used to split a node, the decrease in Gini impurity $\Delta G$ from parent node $t$ to child nodes $t_L$ and $t_R$ (left and right children) is given by:
\begin{equation}
    \label{}
    \Delta G = G(t) - \left( \frac{N_L}{N} G(t_L) + \frac{N_R}{N} G(t_R) \right)
\end{equation}
where $N$ is the total number of samples in the parent node, and $N_L$ and $N_R$ are the numbers of samples in the left and right child nodes, respectively.

The Random Forest calculates the average decrease in Gini impurity across all splits over all trees in the forest for each feature. Features that contribute more significantly to reducing impurity, and thus improving the classification accuracy, are assigned higher importance scores. The feature importance score for feature \( f \) is given by:

\begin{equation}
    \label{}
    \text{Importance}(f) = \frac{1}{N_T} \sum_{t \in T} \Delta G_t(f)
\end{equation}
where $N_T$ is the number of trees in the forest, and $\Delta G_t(f)$ is the decrease in Gini impurity for feature $f$ in tree $t$.

By analyzing the feature importance scores generated by the Random Forest Classifier, we can gain valuable insights into the factors that most influence the demand for space tourism. This information can be critical for informing marketing strategies and resource allocation decisions within the space tourism industry. Focusing on the most impactful features can help companies tailor their offerings and messaging to resonate with potential customers who are more likely to be interested in space travel experiences.

\subsection{Data Augmentation}

Once we had a filtered dataset, data augmentation was applied during model training to promote fairness. This technique creates synthetic data points to balance the dataset, particularly for underrepresented groups. In the context of space tourism, if a specific income bracket or age group has a lower representation in the initial data collection, data augmentation may help address this gap by using the Synthetic Minority Oversampling Technique (SMOTE) \cite{chawla2002smote}. We use SMOTE since it can create new data points for minority classes by interpolating between existing data points within that class. For instance, if the data show a lower number of participants interested in orbital travel compared to suborbital travel, SMOTE can create synthetic data points for the orbital travel category based on existing data points in that group. This helps balance the representation of different travel preferences within the training data for the AI model. By implementing these data cleaning and augmentation techniques, we aim to minimize the influence of biases within the data used to train the AI model. This would ensure that the model makes fairer predictions about the demand for space tourism across various demographics.

\section{Trustworthy and Explainable Artificial Intelligence}

We investigate space tourism demand using the principles of trustworthy and xAI as outlined by the US National Institute of Standards and Technology (NIST) \cite{phillips2021four}. As illustrated in Fig. \ref{TrustworthyAI}, the core of xAI lies in making AI models more understandable and less complex by explaining their inner workings clearly and simply. When a company uses an AI-driven demand prediction model, the model might accurately predict demand for various types of space travel, but company executives need to understand the reasoning behind these predictions to make informed business decisions. For example, the AI model may consider attributes such as price, age, education, and income of potential customers. Similarly, regulatory authorities assessing the safety and feasibility of space travel require explainable AI to ensure transparency and compliance with safety standards. Thus, we developed an xAI method that helps users understand the factors influencing the model's predictions, fostering trust and confidence in its outcomes. In addition to the fairness in data and mitigation of harmful bias described in section \ref{sec.data}, by fostering trust, transparency, and accountability, our xAI method will pave the way for the responsible integration of AI into various domains.

\subsection{Explainability and Interpretability}

Explainability and interpretability are critical components of our trustworthy AI framework, as they enable users to understand how and why the AI model makes specific predictions. By unveiling the rationale behind the model's predictions, we can foster trust and confidence in its outputs. This empowers stakeholders within the space tourism industry to make informed decisions based on not just the predictions themselves, but also the underlying reasoning behind them. In this study, we aim to address several key research questions to better understand the demand for space tourism: 
\begin{itemize}
    \item What kinds of space travel do people like? We investigate the preferences and interests of potential space travelers by analyzing various types of space travel experiences, such as moon travel, suborbital flights, and orbital vacations. By identifying the most popular options, we can tailor offerings to meet market demand.

    \item In each type of travel, what are the distributions of factors? For each identified type of space travel, we analyze the distribution of key factors such as price sensitivity, age demographics, education levels, income brackets, etc. This analysis helps us understand the profile of potential customers for each type of travel experience, enabling more targeted marketing and service design.
    
    \item Given the information about a person, how do we know if this person wants to travel or not? We develop a predictive model that uses individual attributes such as age, income, education level, and interest in space-related activities such as training time, and available tickets, to determine the likelihood of a person wanting to participate in space travel. By making this model explainable, we ensure that the decision-making process is transparent and understandable, allowing stakeholders to trust and effectively utilize the insights generated.
\end{itemize}
By answering these questions, our study provides a comprehensive understanding of space tourism demand, leveraging trustworthy and explainable AI to make informed, unbiased, and strategic decisions in this emerging field.

One powerful approach we leverage for achieving interpretability is the SHapley Additive exPlanations (SHAP) technique \cite{lundberg2017unified}. SHAP assigns an attribution value, a SHAP value, to each feature for a given prediction. This value represents the contribution of that specific feature to the model's final output. High positive SHAP values indicate that the feature has a strong positive influence on the prediction. Conversely, low negative SHAP values signify a negative influence. By analyzing the SHAP values for each feature, we gain insights into the relative importance of various factors shaping the model's predictions about space tourism demand.

Utilizing SHAP in our space tourism demand prediction model allows us to obtain several benefits and answer the questions that we mentioned above. First, it enables us to identify the key drivers that most significantly influence the model's predictions. This knowledge empowers space tourism companies to prioritize factors that resonate most with potential customers. For instance, a high SHAP value for "moon travel" would suggest this experience is particularly appealing to a significant portion of potential customers. Conversely, low negative SHAP values signify a negative influence.

We have four travel options in our study, by using SHAP we can specify the impact of each feature to each option. Furthermore, in each option, we can visualize and quantify the impact of each feature on the model's predictions. This helps in understanding the key drivers behind the demand for space tourism and in making the AI model's decision process transparent. For example, SHAP values can reveal whether income, age, or education level have a more significant influence on the likelihood of an individual being interested in space travel. This information is invaluable for tailoring marketing strategies and making informed business decisions.

\subsection{Validity and Reliability}

The trustworthiness of our space tourism demand prediction model hinges on its validity and reliability. Validity refers to the model's ability to generate accurate and relevant predictions, while reliability concerns the consistency and stability of these predictions over time. We employ a multi-pronged approach to guarantee the validity of our model. First, we leverage k-fold cross-validation (k = 5). This technique partitions the data into k folds, trains the model on k-1 folds, and assesses its performance on the remaining folds. This process iterates k times, ensuring the model generalizes well to unseen data. 

To further solidify the model's validity, we performed an external validation. This involves testing the model on an independent dataset that mirrors the size, distribution, and real-world characteristics of the training data. By demonstrating strong performance in this external environment, we strengthened confidence that the model generalizes beyond the training data and accurately reflects the real-world space tourism demand. Additionally, we evaluated the model's performance using the Area Under the Curve (AUC) metric \cite{metz1978basic}. AUC provides a comprehensive view of the model's ability to distinguish between potential space tourists and those with no interest. We will discuss the metric in detail in Section \ref{sec.evaluationMetric}. Furthermore, we employ test-retest reliability to assess the model's performance over time. This involves evaluating the model on the same data points at different time intervals. Consistent predictions across these evaluations indicate reliable model behavior, signifying that the model's predictions remain stable over time.

By validating and ensuring the reliability of our AI model, we elevate its trustworthiness. This translates into the generation of accurate and consistent predictions, a critical factor for informed decision-making in the space tourism industry. Given the high stakes involved, reliable insight from the model is paramount for success, influencing both business strategy and customer satisfaction. 

\section{Transformer Neural Network}
\label{sec.transformer_model}

\begin{figure*}[h]
    \centering
    \includegraphics[width=0.85\linewidth]{ 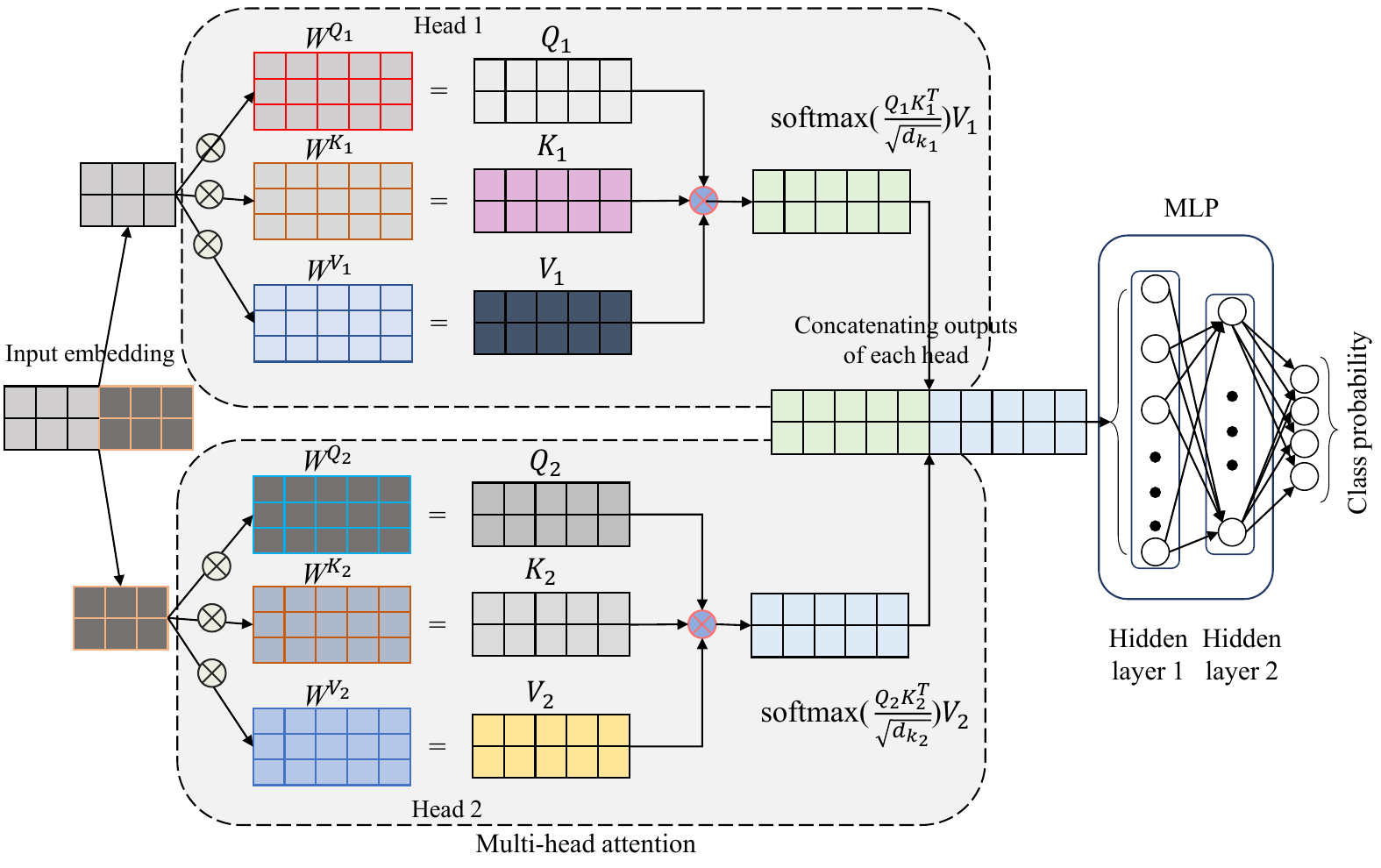}
    \caption{Tranformer network.}
    \label{fig.transformernetwork}
\end{figure*}

We developed a novel tourism prediction model, named SpaceNet, by leveraging transformer neural networks \cite{vaswani2017attention}. The SpaceNet architecture involves processing sequential data, making them particularly well-suited for datasets with numerous features, and for analyzing factors that might influence an individual's interest in space travel. The architecture of our SpaceNet model is shown in Fig. \ref{fig.transformernetwork}. Unlike traditional neural networks such as RNNs and LSTM, transformer-based models can effectively capture long-range dependencies in the data. This capability is crucial for understanding complex relationships between various factors that might influence someone's decision to participate in space tourism. For instance, a person's age might influence their risk tolerance, consequently influencing their preference for a specific type of travel option (e.g., orbital or moon). In this context, capturing these long-range interactions becomes imperative for precise demand prediction.

The SpaceNet is composed of several key components that work together to analyze the data and generate predictions including the embedding layer, multi-layer transformer, and multi-layer perception (MLP). As illustrated in Fig. \ref{fig.transformernetwork}, our model sequentially processes input data by passing an embedding layer, multi-layer transformer, and fully connected layers before outputting a prediction. Our specific implementation utilizes a multi-layer transformer with two encoder layers. Each layer employs a hidden size of sixty-four and two attention heads in the multi-head attention mechanism. We used this configuration after considering factors such as the complexity of the relationships within our data, the size of our dataset, and the obtained results. This configuration allows the model to learn intricate relationships between features as well as avoid overfitting while maintaining computational efficiency.

In the \textbf{embedding layer}, the input data including multiple individual attributes is transformed into a higher-dimensional vector space. This process allows the model to represent complex relationships between features more effectively. Sequentially, this vector is processed by the \textbf{multi-head attention}, the core of the model, which consists of multiple stacked layers. Multi-head attention is a key component within the transformer encoder. It essentially allows the model to attend to, or focus on, specific parts of the input sequence when encoding information. 

There are two attention heads in our model, focusing on relevant portions of the input sequence and capturing long-range dependencies between features. In an attention head, the input data is first projected into three different vector spaces using separate weight matrices $W^Q$, $W^V$, and $W^K$. These projected vectors are called query ($Q$), key ($K$), and value ($V$) vectors. Intuitively, the query vector represents what we are interested in focusing on, the key vector represents where to find relevant information in the sequence, and the value vector holds the actual information from those relevant parts.

The query vectors are then compared to all the key vectors using a dot product operation ($QK^T$). To prevent exploding gradients during training, these dot products are then scaled by a constant value as the square root of the dimension of the key vectors ($d_k$). This scaled score indicates how well a particular key vector (a specific data point related to age) aligns with the current query (our current focus on age). Sequentially, a softmax function is applied to the scaled dot-product scores, converting them into attention weights. Finally, the attention weights are multiplied element-wise with the value vectors, which contain the actual information from each position. This effectively creates a context vector that summarizes the most relevant information from the entire input sequence, considering the current focus of the query vector.

By repeating this self-attention process across multiple heads within the multi-head attention mechanism, the model can learn from different aspects of the data simultaneously. This allows for a richer understanding of the complex relationships between various factors influencing space tourism interest. For example, the model might attend to age and risk tolerance simultaneously through separate heads, ultimately capturing how these various factors are combined to influence travel preference (e.g. moon vs. orbital).

Following the multi-head attention, an MLP with two fully connected layers transforms the multi-head attention output into a lower-dimension vector and then outputs the desired output size. This output vector typically represents the probability of classes or travel options. In addition, we add a dropout layer followed by the first fully connected layer to prevent overfitting. This technique randomly drops a certain percentage of neurons during each training iteration, forcing the model to learn robust features that are not overly reliant on specific data points.

By leveraging the strengths of the transformer architecture, our model can effectively capture the complex relationships within the space tourism demand data and generate accurate predictions about potential customer interest. This approach allows us to not only predict demand but also gain insights into the factors that most significantly influence an individual's decision to participate in space travel.

\begin{figure*}[h]
    \centering
    \begin{subfigure}{0.495\textwidth}
        \centering
        \includegraphics[width=\textwidth]{ 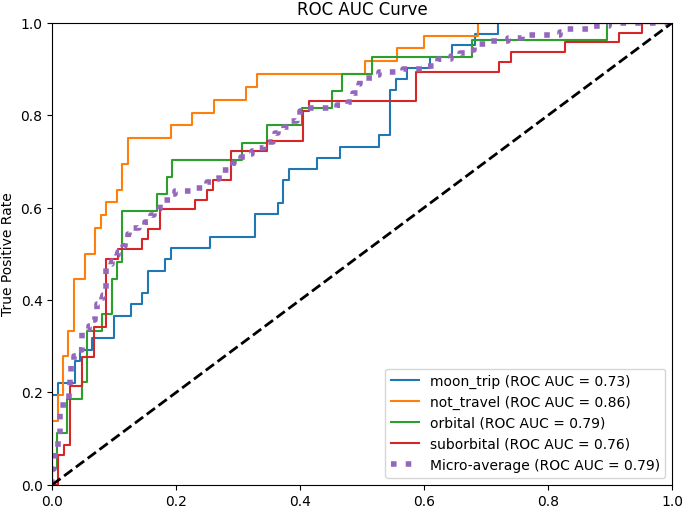}
        \caption{Roc AUC using all features.}
        \label{fig.allfeature}
    \end{subfigure}
    \begin{subfigure}{0.495\textwidth}
        \centering
        \includegraphics[width=\textwidth]{ 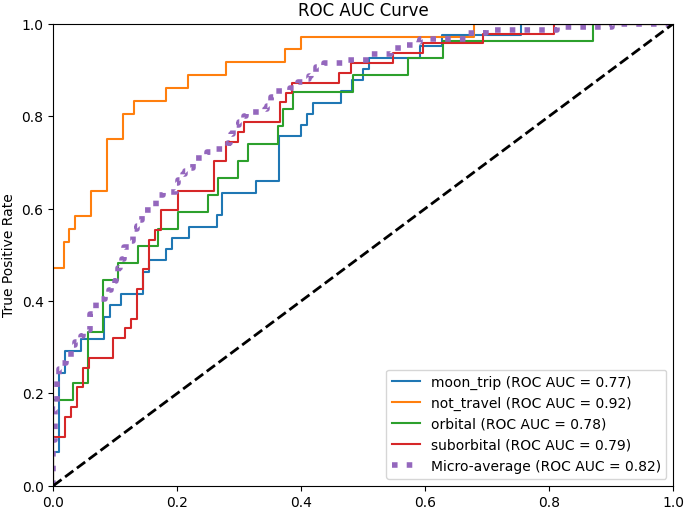} 
        \caption{Roc AUC using the top 25 features.}
        \label{fig.top25feature}
    \end{subfigure}
   
    \caption{Comparison of ROC AUC when using all features and top 25 features.}
    \label{fig.ROCAUC_prediction}
\end{figure*}

\section{Experiment and Evaluation Metrics}
\subsection{Model Training}
\label{sec.modelTraining}

Building upon the custom transformer neural network architecture detailed in Section~\ref{sec.transformer_model}, we implemented our space tourism demand prediction model using PyTorch 1.13.1, a deep learning framework. To accelerate training for this computationally intensive model, we leveraged the AI.Panther high-performance computing facility hosted at the Florida Institute of Technology. Equipped with the A100 SXM4 GPUs, this facility significantly accelerated deep-learning training compared to conventional hardware.

During training, we employed the Adam optimizer with a learning rate of 0.001 and weight decay of 1e-5 to promote model stability and prevent overfitting. Weight decay penalizes large weights while encouraging the model to learn more robust features that generalize better to unseen data. We balanced training efficiency and model complexity by configuring the process with a batch size of 64 and a total of 50 epochs. The cross-entropy loss function served as the loss function, guiding optimization by measuring the discrepancy between the model's predictions and the ground truth labels.

To achieve a multifaceted understanding of our model's performance and feature influence, we conducted experiments using two distinct feature sets: Fully collected features and top-25 important features. For the full feature set, the first experiment employed the entire set of features present in the original dataset. This comprehensive approach allows us to assess the model's ability to leverage all available information for accurate space tourism demand prediction. It provides a baseline understanding of the model's overall performance without introducing potential biases through feature selection.

For the top-25 feature set, we focused on a reduced set containing the top-25 most important features identified through the random forest classification algorithm as mentioned in section \ref{subsec.data_cleaning}. This approach streamlines the training process and computational requirements. More importantly, it allows us to isolate the impact of the most influential features on the model's predictions. By analyzing the model's performance with this reduced set, we can gain insights into whether the model can achieve comparable accuracy while relying on a smaller, more targeted set of features. This knowledge can be valuable for real-world applications where space travel companies can save their time by investigating mainly these features. It also provides insights into niche tourism markets to target specific tourist segments and guide sustainable tourism practices.

\subsection{Evaluation Metrics}
\label{sec.evaluationMetric}
We use the ROC AUC metric to assess the model's ability to discriminate between potential space tourists and those with no interest \cite{metz1978basic}. Unlike accuracy, which can be misleading in imbalanced datasets (common in space tourism prediction when the number of traveled people is significantly less than non-traveled people), ROC AUC offers a robust and comprehensive evaluation metric. It considers both the True Positive Rate (TPR), representing correctly identified space tourists, and the False Positive Rate (FPR), indicating incorrectly classified non-tourists. Mathematically, ROC AUC can be expressed as the integral of the ROC curve, which plots TPR against FPR across all possible classification thresholds:

\begin{equation}
    \label{roc_auc}
    \text{ROC AUC} = \int_{0}^{1} \text{TPR}(\tau) \, \text{dFPR}(\tau).
\end{equation}
In general, a higher AUC indicates better performance, and AUC values from 0.8 to 0.9 illustrate good model prediction.

The ROC curve itself visualizes this trade-off between TPR and FPR. Formally, FPR is defined as the ratio of False Positives (FP) to the sum of FP and True Negatives (TN), while TPR represents the proportion of True Positives (TP) to the sum of TP and False Negatives (FN):
\begin{align}
    \label{FPR}
    \text{FPR} &= \frac{\text{FP}}{\text{FP} + \text{TN}} \mbox{ and}\\
    \label{TPR}
    \text{TPR} &= \frac{\text{TP}}{\text{TP} + \text{FN}}
\end{align}
where TP, FP, TN, and FN denote true positives, false positives, true negatives, and false negatives, respectively. An AUC of 1 signifies a perfect model, flawlessly distinguishing space tourism interest, while an AUC of 0.5 reflects performance no better than random guessing. A key advantage of ROC AUC is its indifference to class imbalance, making it a reliable measure of model performance across varying thresholds. For multi-class classification problems, like ours, ROC AUC can be extended using methods such as One-vs-One (OvO) or One-vs-Rest (OvR) to assess the model's ability to distinguish between all classes.

\begin{figure*}[h]
    \centering
    \includegraphics[width=0.8\linewidth]{ 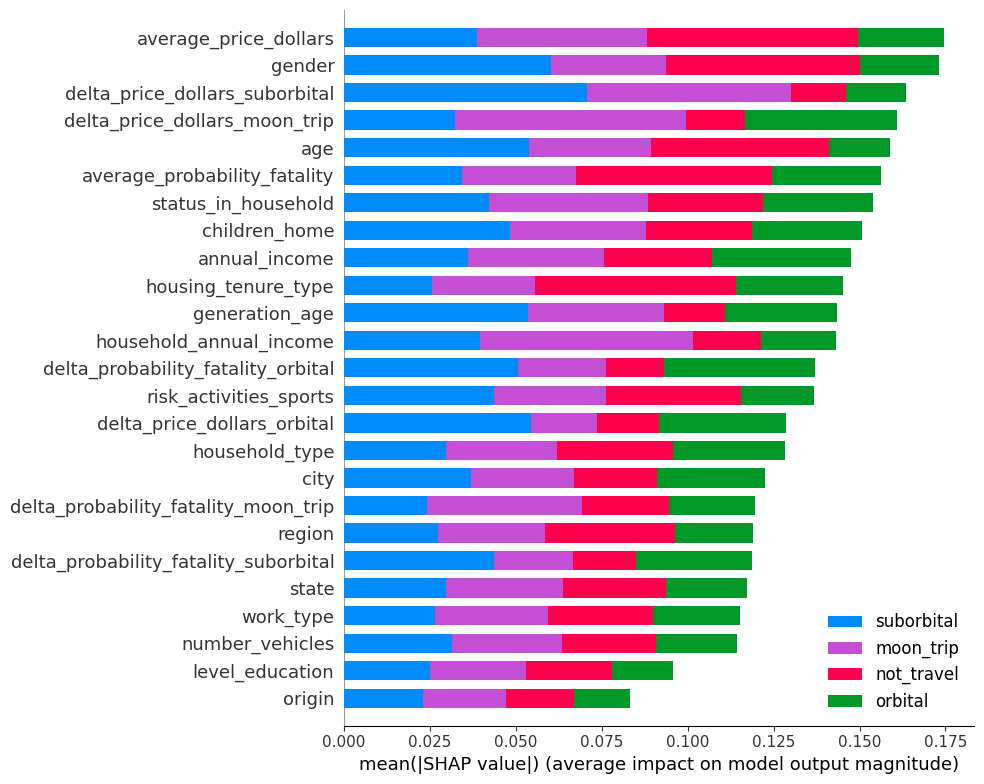}
    \caption{Global feature contribution.}
    \label{fig.globalsummary_plot}
\end{figure*}

\begin{figure*}[h]
    \centering
    \small
    \begin{subfigure}{0.385\textwidth}
        \centering
        \includegraphics[width=\textwidth]{ 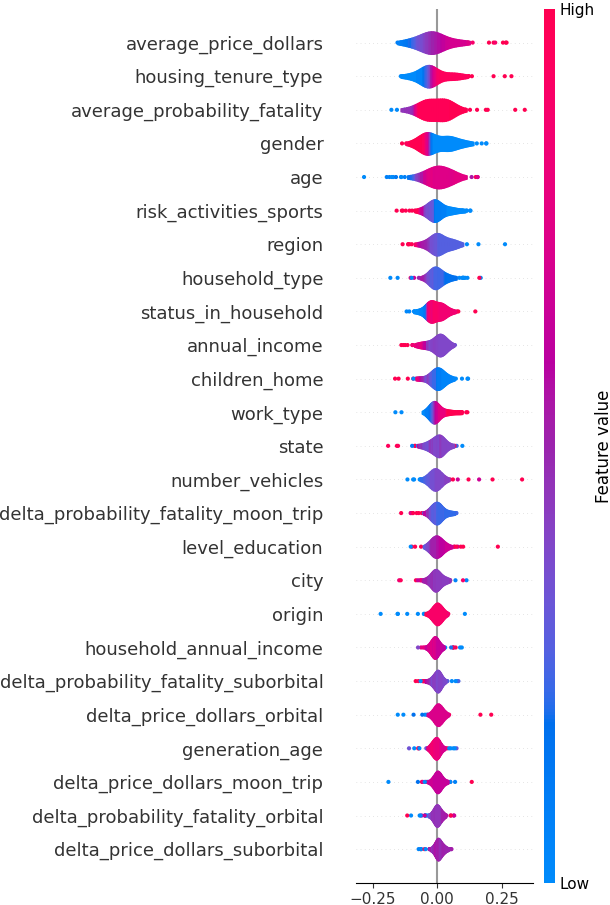}
        \caption{No travel.}
        \label{fig.trave}
    \end{subfigure}
    \begin{subfigure}{0.385\textwidth}
        \centering
        \includegraphics[width=\textwidth]{ 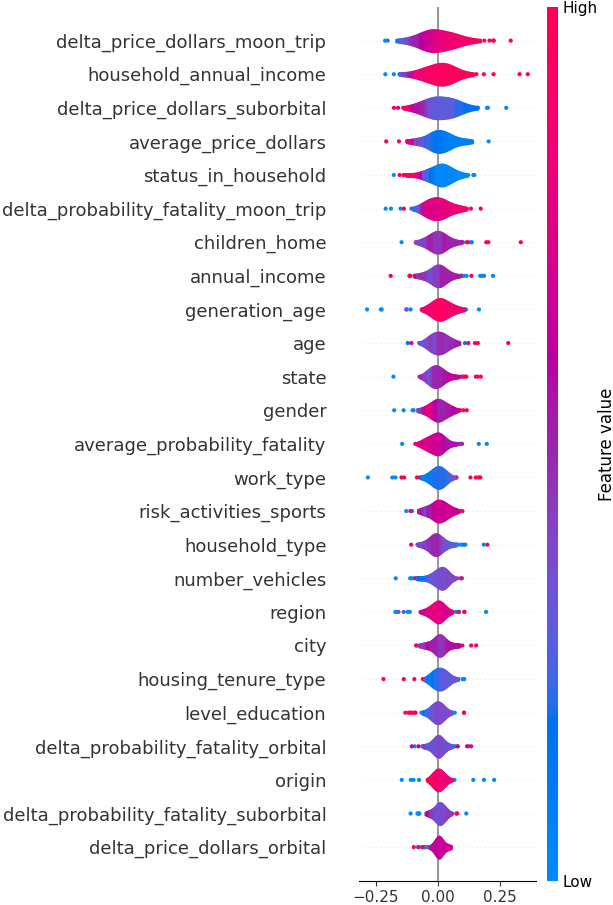}
        \caption{Moon trip.}
        \label{fig.moontrip}
    \end{subfigure}
    \vspace{1em} 
    \begin{subfigure}{0.385\textwidth}
        \centering
        \includegraphics[width=\textwidth]{ 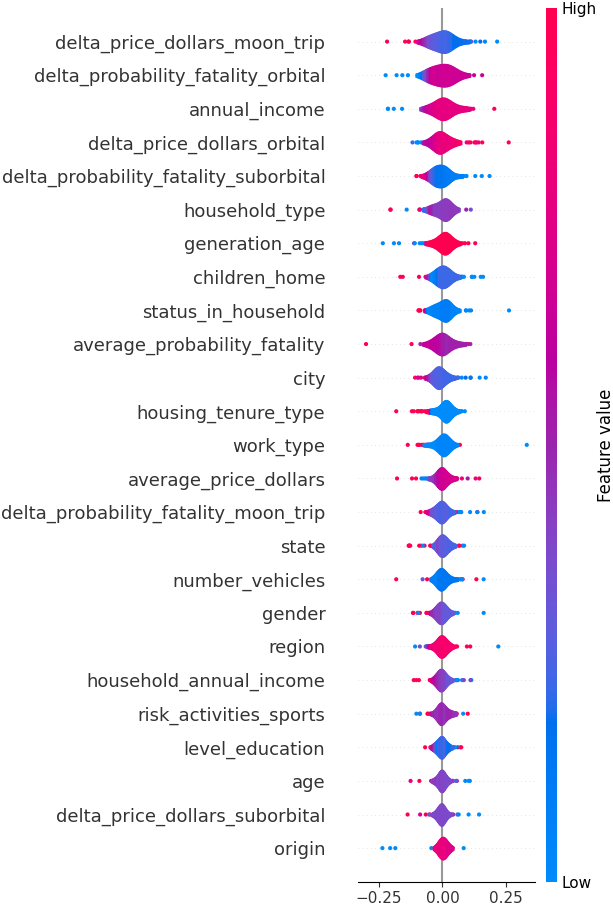}
        \caption{Suborbital travel.}
        \label{fig.suborbital}
    \end{subfigure}
    \begin{subfigure}{0.385\textwidth}
        \centering
        \includegraphics[width=\textwidth]{ 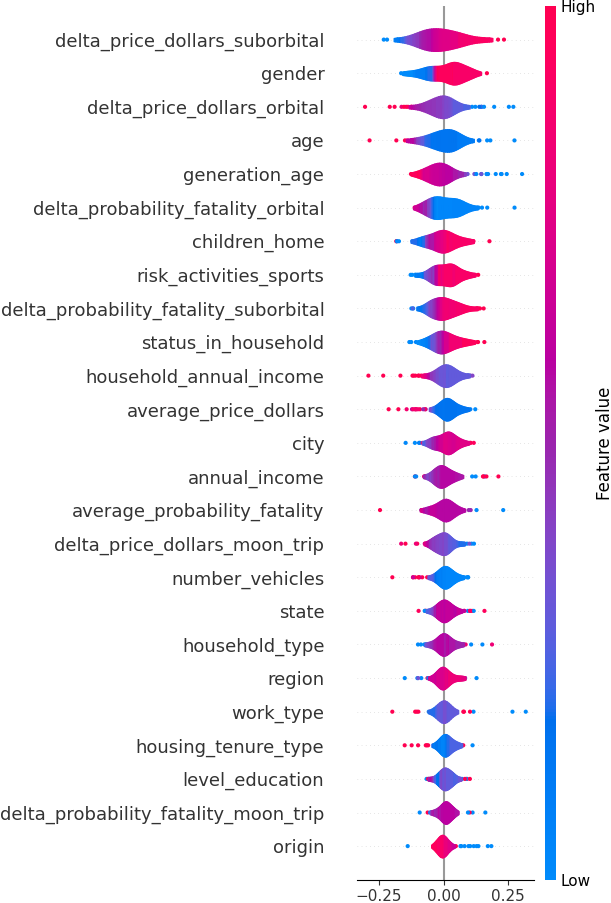}
        \caption{Orbital travel.}
        \label{fig.orbital}
    \end{subfigure}
    \caption{Explanation for individual classes.}
    \label{fig.individual_classes_prediction}
\end{figure*}

\section{Result and Discussion}
\label{sec.analysisAndDiscussion}
        
\subsection{Travel Demand Using 5-folds Cross-validation Training Method}

The experimental result is illustrated in Fig. \ref{fig.ROCAUC_prediction}, which shows the classification performance when using all features (Fig. \ref{fig.allfeature}) and when using the top-25 features (Fig. \ref{fig.top25feature}). In particular, our model achieved an ROC AUC of 0.79 $\pm$ 0.067 on the entire dataset, indicating a good performance in predicting space tourism demand. For the top 25 features, the results outperformed the full set with an ROC AUC of 0.82 $\pm$ 0.088, which indicates our SpaceNet performance in space travel prediction. This suggests that the model can achieve comparable or even better accuracy using a smaller set of the most influential features. As a result, feature selection using the top 25 most important features can improve the model's performance, potentially due to reduced noise or redundancy in the full feature set.

\subsection{Explainable AI}

In this section, we focus on analyzing and discussing our space tourism demand prediction model's key findings, with a particular focus on the principles of xAI. We dissect the model's performance through three key explanation techniques: 
\begin{itemize}
    \item Global explanation: Investigation of the overall important features.
    \item Local explanation: Investigation of the feature's contribution to individual space travel prediction.
    \item Instance explanation: Investigation of the reasoning behind a data point prediction.
\end{itemize}
By employing a multifaceted approach to explainability, we not only assess the model's accuracy but also gain valuable insights into the reasoning behind its predictions. This fosters trust in the model's decision-making process and ultimately empowers stakeholders within the space tourism industry.

\subsubsection{Global Explanation}
\label{sec.globalExplanation}

We leverage SHAP techniques to explain the impact of each feature on the model's prediction for a given data point. By calculating the average absolute SHAP value for each feature across the entire dataset, we can gain insights into the overall feature importance for space tourism demand prediction. The visual representation of these average absolute SHAP values is depicted in Fig.~\ref{fig.globalsummary_plot}, in which features are listed on the y-axis, and the absolute SHAP values (higher values indicate greater impact on the model's output) are on the x-axis. For each feature, the distribution of each travel class - non-travel, suborbital, orbital, and moon travel - is highlighted in red, cyan, green, and purple, respectively.

In the experiment, the feature ``average\_price\_dollars" appears to have the greatest impact on the model’s output, followed by ``age" and ``delta\_price\_dollars\_moon\_trip". This finding aligns with expectations, as price and age are factors commonly considered in decision-making. It is important to note that the order of importance for these features might vary slightly depending on the initial weights assigned during model training, although feature rankings generally fluctuate within a range of $\pm$3. For example, the SHAP values for ``annual\_income" and ``gender" are relatively close, suggesting their order could potentially swap with different initial weight configurations.

\subsubsection{Local Explanation}
\label{sec.LacalExplanation}

Local explanations take us a step further to understand predictions, which provide insights into how specific features contribute to individual predictions. This is particularly valuable for comprehending why the model makes certain predictions for a particular class (space travel options). Here, we will explore how the model leverages different features to predict space tourism interest for an individual class, as shown in Fig. \ref{fig.individual_classes_prediction}. Again, features are shown on the left side of each subfigure following an important order from the top to bottom. The red color represents high values, and the blue color illustrates low values.

\begin{figure*}[h]
    \centering
    \includegraphics[width=\linewidth]{ 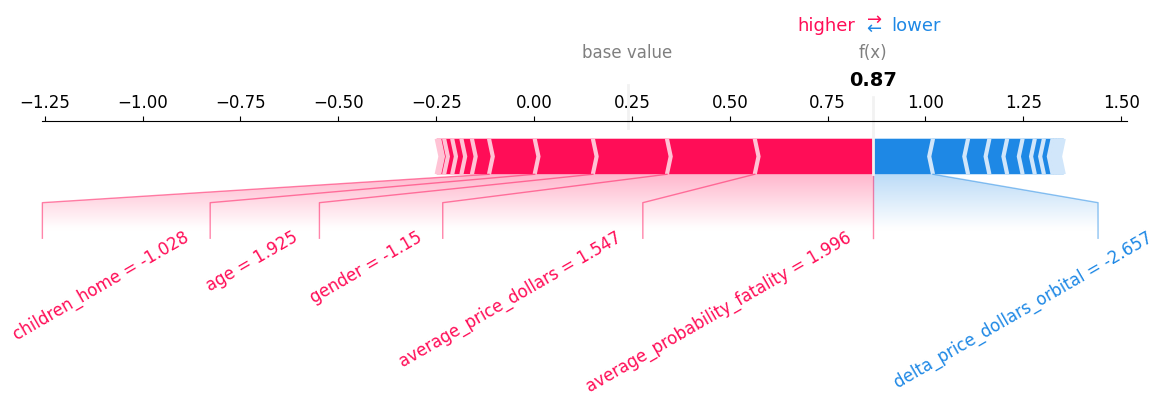}
    \caption{Instance explanation.}
    \label{fig.fig_instance_explanation}
\end{figure*}

From our investigation, the high ticket price makes people not want to travel. Young people tend to travel to space rather than old people, as illustrated in Fig. \ref{fig.trave}. Males (red color) want to travel more than females. People who have lower annual income do not want to travel. High fatality probability makes people do not want to travel. The trend of these features is mostly converse in the travel classes as illustrated in Fig \ref{fig.moontrip}, Fig. \ref{fig.suborbital}, and Fig. \ref{fig.orbital}. This result indicates that the prediction of our SpaceNet aligns well with natural human tendencies, emphasizing the robustness and reliability of our model in space travel demand prediction.

\subsubsection{Instance Explanation}
\label{sec.instanceExplantion}

To understand why a particular individual may or may not choose to travel, we use instance explanations to justify a model’s prediction for a single data point, as shown in Fig. \ref{fig.fig_instance_explanation}. This approach enhances model fairness and transparency. In this case, the model predicts a travel likelihood of 0.87 for an individual. The key positive factors contributing to this prediction include the number of children at home, age, gender, ticket price, and probability of fatality, while the primary negative factor is the cost of orbital travel. Each attribute value is scaled using standard normalization, as described in Section \ref{subsec.data_cleaning}. The actual corresponding values are as follows: number of children = 0, age = 65, gender = female, average ticket price = 65,547, and average fatality probability = 7.5\%.

This instance explanation aligns with broader trends discussed in Sections \ref{sec.globalExplanation} and \ref{sec.LacalExplanation}, which highlights age, ticket price, and fatality risk as key factors influencing travel decisions. By providing a clear rationale for the model’s prediction, instance explanations foster trust and support informed decision-making in the space tourism industry.

The explainability of our SpaceNet framework is crucial as it makes the complex deep-learning system and its decisions understandable and interpretable to humans. This is especially important for building trust and providing explanations for how the SpaceNet model reaches its conclusions. Our SpaceNet model ensures accountability, reduces biases, and helps users verify fairness, accuracy, and safety in a transparent manner.

\section{Conclusion}
\label{sec.conclusion}

This paper proposes the novel deep-learning SpaceNet model for predicting the space tourism demand of people living in the US. The framework is developed with an explainability pipeline for trustworthy and reliable AI decision-making. In particular, we prioritized model interpretability by following to National Institute of Standards and Technology guidelines for trustworthy AI and incorporating Shapley additive explanations techniques. 

In our experiments, SpaceNet achieved impressive results in predicting space travel types (no travel, moon travel, suborbital, orbital) with an ROC AUC of 0.82 $\pm$ 0.088. This demonstrates the model's strong predictive performance and its generalizability across diverse demographics. Beyond travel forecasting, our approach delves deeper into the key factors influencing space travel decisions. From the experiment, we identify travel price, age, income, and gender are important features among many factors. This granular understanding of the ``why" behind travel choices provides a significant advantage for space tourism businesses. By leveraging the insights extracted from our SpaceNet model and the proposed xAI pipeline, companies can gain a deeper understanding of their customers' demands. This knowledge empowers them to tailor marketing strategies with optimized service offerings to cater to specific demographics. Furthermore, this research lays the groundwork for exciting future exploration in space tourism demand prediction. 

While this study demonstrates the effectiveness of our deep learning approach for space tourism demand prediction, there are limitations to consider for follow-up work such as a diversity of nationality. Future research could also delve into segmenting the target audience based on travel type preferences such as moon travel to provide even more actionable insights for space tourism businesses. By doing that, we can shape the future of the space tourism industry by enabling data-driven strategies that cater to the specific desires and motivations of potential space travelers. 

\section*{CRediT Authorship Contribution Statement}
\textbf{Tan-Hanh Pham}: Conceptualization, Investigation, Methodology, Formal analysis, Validation, Visualization, Writing - original draft, Writing – review \& editing. \textbf{Jingchen Bi}: Data Collection, Formal analysis, Methodology, Writing - original draft. \textbf{Rodrigo Mesa-Arango}: Conceptualization, Methodology, Data Collection, Formal analysis, Writing – review \& editing. \textbf{Kim-Doang Nguyen}: Conceptualization, Methodology, Formal analysis, Writing – review \& editing, Funding acquisition.

\section*{Acknowledgment}

This material is based upon work supported by the U.S. National Science Foundation under Grant \#2245022.

\balance
{\small
\bibliographystyle{ieee_fullname}
\bibliography{egbib}

\begin{thebibliography}{10}\itemsep=-1pt

\bibitem{achiam2023gpt}
Josh Achiam, Steven Adler, Sandhini Agarwal, Lama Ahmad, Ilge Akkaya, Florencia~Leoni Aleman, Diogo Almeida, Janko Altenschmidt, Sam Altman, Shyamal Anadkat, et~al.
\newblock Gpt-4 technical report.
\newblock {\em arXiv preprint arXiv:2303.08774}, 2023.

\bibitem{breiman2001random}
Leo Breiman.
\newblock Random forests.
\newblock {\em Machine learning}, 45:5--32, 2001.

\bibitem{burgess2015friendship}
Colin Burgess.
\newblock {\em Friendship 7: The Epic Orbital Flight of John H. Glenn, Jr.}
\newblock Springer, 2015.

\bibitem{burgess2022soviets}
Colin Burgess.
\newblock {\em Soviets in Space: Russia’s Cosmonauts and the Space Frontier}.
\newblock Reaktion books, 2022.

\bibitem{chang2017preliminary}
Yi-Wei Chang.
\newblock A preliminary examination of the relationship between consumer attitude towards space travel and the development of innovative space tourism technology.
\newblock {\em Current Issues in Tourism}, 20(14):1431--1453, 2017.

\bibitem{chawla2002smote}
Nitesh~V Chawla, Kevin~W Bowyer, Lawrence~O Hall, and W~Philip Kegelmeyer.
\newblock Smote: synthetic minority over-sampling technique.
\newblock {\em Journal of artificial intelligence research}, 16:321--357, 2002.

\bibitem{crouch2009modelling}
Geoffrey~I Crouch, Timothy~M Devinney, Jordan~J Louviere, and Towhidul Islam.
\newblock Modelling consumer choice behaviour in space tourism.
\newblock {\em Tourism Management}, 30(3):441--454, 2009.

\bibitem{crouch2003bishop}
Tom~D Crouch.
\newblock {\em The bishop's boys: A life of Wilbur and Orville Wright}.
\newblock WW Norton \& Company, 2003.

\bibitem{davazdahemami2022explanatory}
Behrooz Davazdahemami, Hamed~M Zolbanin, and Dursun Delen.
\newblock An explanatory machine learning framework for studying pandemics: The case of covid-19 emergency department readmissions.
\newblock {\em Decision Support Systems}, 161:113730, 2022.

\bibitem{dekking2005modern}
Frederik~Michel Dekking.
\newblock {\em A Modern Introduction to Probability and Statistics: Understanding why and how}.
\newblock Springer Science \& Business Media, 2005.

\bibitem{gatti2023assessing}
Matteo Gatti, Irene Ceccato, Adolfo Di~Crosta, Pasquale La~Malva, Emanuela Bartolini, Rocco Palumbo, Alberto Di~Domenico, and Nicola Mammarella.
\newblock Assessing space tourism propensity: A new questionnaire for future space tourists.
\newblock {\em Aerospace}, 10(12):1018, 2023.

\bibitem{blueorigin}
Will Gendron.
\newblock Virgin galactic flight - players in space tourism, 2024.

\bibitem{Start}
Sarah Hoffschwelle.
\newblock Space tourism: How much does it cost \& who's offering it?, 2024.

\bibitem{kim2023space}
Myung~Ja Kim, C~Michael Hall, and Ohbyung Kwon.
\newblock Space tourism: Do age and gender make a difference in risk perception?
\newblock {\em Journal of Hospitality and Tourism Management}, 57:13--17, 2023.

\bibitem{kim2024orbital}
Myung~Ja Kim, Colin~Michael Hall, Ohbyung Kwon, Kyunghwa Hwang, and Jinok~Susanna Kim.
\newblock Orbital and sub-orbital space tourism: motivation, constraint and artificial intelligence.
\newblock {\em Tourism Review}, 79(2):392--407, 2024.

\bibitem{knapivc2021explainable}
Samanta Knapi{\v{c}}, Avleen Malhi, Rohit Saluja, and Kary Fr{\"a}mling.
\newblock Explainable artificial intelligence for human decision support system in the medical domain.
\newblock {\em Machine Learning and Knowledge Extraction}, 3(3):740--770, 2021.

\bibitem{laing2004australian}
Jennifer~H Laing and Geoffrey~I Crouch.
\newblock Australian public interest in space tourism and a cross-cultural comparison.
\newblock {\em Journal of Tourism Studies}, 15(2):26--36, 2004.

\bibitem{laing2005extraordinary}
Jennifer~H Laing and Geoffrey~I Crouch.
\newblock Extraordinary journeys: An exploratory cross-cultural study of tourists on the frontier.
\newblock {\em Journal of Vacation Marketing}, 11(3):209--223, 2005.

\bibitem{le2013astrium}
Thierry Le~Goff and Antoine Moreau.
\newblock Astrium suborbital spaceplane project: Demand analysis of suborbital space tourism.
\newblock {\em Acta Astronautica}, 92(2):144--149, 2013.

\bibitem{lecun2015deep}
Yann LeCun, Yoshua Bengio, and Geoffrey Hinton.
\newblock Deep learning.
\newblock {\em nature}, 521(7553):436--444, 2015.

\bibitem{lindenmoyer2015commercial}
Alan Lindenmoyer, Mike Horkachuck, Gwynne Shotwell, Bruce Manners, and Frank Culbertson.
\newblock Commercial orbital transportation services (cots) program lessons learned.
\newblock Technical report, 2015.

\bibitem{lundberg2017unified}
Scott~M Lundberg and Su-In Lee.
\newblock A unified approach to interpreting model predictions.
\newblock {\em Advances in neural information processing systems}, 30, 2017.

\bibitem{mazzucato2018co}
Mariana Mazzucato and Douglas~KR Robinson.
\newblock Co-creating and directing innovation ecosystems? nasa's changing approach to public-private partnerships in low-earth orbit.
\newblock {\em Technological Forecasting and Social Change}, 136:166--177, 2018.

\bibitem{mesa2023missions}
Rodrigo Mesa-Arango, Juan Pineda-Jaramillo, Diogo~SA Araujo, Jingchen Bi, Mahesh Basva, and Francesco Viti.
\newblock Missions and factors determining the demand for affordable mass space tourism in the united states: A machine learning approach.
\newblock {\em Acta Astronautica}, 204:307--320, 2023.

\bibitem{metz1978basic}
Charles~E Metz.
\newblock Basic principles of roc analysis.
\newblock In {\em Seminars in nuclear medicine}, volume~8, pages 283--298. Elsevier, 1978.

\bibitem{neufeld2018spaceflight}
Michael~J Neufeld.
\newblock {\em Spaceflight: a concise history}.
\newblock MIT Press, 2018.

\bibitem{nimmy2022explainability}
Sonia~Farhana Nimmy, Omar~K Hussain, Ripon~K Chakrabortty, Farookh~Khadeer Hussain, and Morteza Saberi.
\newblock Explainability in supply chain operational risk management: A systematic literature review.
\newblock {\em Knowledge-Based Systems}, 235:107587, 2022.

\bibitem{nordin2023explainable}
Noratikah Nordin, Zurinahni Zainol, Mohd Halim~Mohd Noor, and Lai~Fong Chan.
\newblock An explainable predictive model for suicide attempt risk using an ensemble learning and shapley additive explanations (shap) approach.
\newblock {\em Asian journal of psychiatry}, 79:103316, 2023.

\bibitem{olya2023emerging}
Hossein Olya and Heesup Han.
\newblock Emerging space tourism business: Uncovering customer avoidance responses and behaviours.
\newblock {\em Journal of Vacation Marketing}, 29(3):445--460, 2023.

\bibitem{olya2020antecedents}
Hossein~GT Olya and Heesup Han.
\newblock Antecedents of space traveler behavioral intention.
\newblock {\em Journal of Travel Research}, 59(3):528--544, 2020.

\bibitem{pham2024deep}
Tan-Hanh Pham, Praneel Acharya, Sravanthi Bachina, Kristopher Osterloh, and Kim-Doang Nguyen.
\newblock Deep-learning framework for optimal selection of soil sampling sites.
\newblock {\em Computers and Electronics in Agriculture}, 217:108650, 2024.

\bibitem{pham2024adaptive}
Tan-Hanh Pham, Godwyll Aikins, Tri Truong, and Kim-Doang Nguyen.
\newblock Adaptive compensation for robotic joint failures using partially observable reinforcement learning.
\newblock {\em Algorithms}, 17(10):436, 2024.

\bibitem{pham2025enhanced}
Tan-Hanh Pham, Travis Burgers, and Kim-Doang Nguyen.
\newblock Enhanced droplet analysis using generative adversarial networks.
\newblock {\em Computers and Electronics in Agriculture}, 231:109922, 2025.

\bibitem{pham2023seunet}
Tan-Hanh Pham, Xianqi Li, and Kim-Doang Nguyen.
\newblock Seunet-trans: A simple yet effective unet-transformer model for medical image segmentation.
\newblock {\em IEEE Access}, 2024.

\bibitem{pham2024soil}
Tan-Hanh Pham and Kim-Doang Nguyen.
\newblock Soil sampling map optimization with a dual deep learning framework.
\newblock {\em Machine Learning and Knowledge Extraction}, 6(2):751--769, 2024.

\bibitem{phillips2021four}
P~Jonathon Phillips, P~Jonathon Phillips, Carina~A Hahn, Peter~C Fontana, Amy~N Yates, Kristen Greene, David~A Broniatowski, and Mark~A Przybocki.
\newblock Four principles of explainable artificial intelligence.
\newblock 2021.

\bibitem{reddy2012space}
Maharaj~Vijay Reddy, Mirela Nica, and Keith Wilkes.
\newblock Space tourism: Research recommendations for the future of the industry and perspectives of potential participants.
\newblock {\em Tourism Management}, 33(5):1093--1102, 2012.

\bibitem{sariyer2024leveraging}
Gorkem Sariyer, Sachin~Kumar Mangla, Mert~Erkan Sozen, Guo Li, and Yigit Kazancoglu.
\newblock Leveraging explainable artificial intelligence in understanding public transportation usage rates for sustainable development.
\newblock {\em Omega}, 127:103105, 2024.

\bibitem{nasa_airplanes}
Robert~J. Shaw.
\newblock Ultra-efficient engine technology (ueet) program, airplanes, 2024.

\bibitem{shrestha2019review}
Ajay Shrestha and Ausif Mahmood.
\newblock Review of deep learning algorithms and architectures.
\newblock {\em IEEE access}, 7:53040--53065, 2019.

\bibitem{team2023gemini}
Gemini Team, Rohan Anil, Sebastian Borgeaud, Yonghui Wu, Jean-Baptiste Alayrac, Jiahui Yu, Radu Soricut, Johan Schalkwyk, Andrew~M Dai, Anja Hauth, et~al.
\newblock Gemini: a family of highly capable multimodal models.
\newblock {\em arXiv preprint arXiv:2312.11805}, 2023.

\bibitem{touvron2023llama}
Hugo Touvron, Thibaut Lavril, Gautier Izacard, Xavier Martinet, Marie-Anne Lachaux, Timoth{\'e}e Lacroix, Baptiste Rozi{\`e}re, Naman Goyal, Eric Hambro, Faisal Azhar, et~al.
\newblock Llama: Open and efficient foundation language models.
\newblock {\em arXiv preprint arXiv:2302.13971}, 2023.

\bibitem{vaswani2017attention}
Ashish Vaswani, Noam Shazeer, Niki Parmar, Jakob Uszkoreit, Llion Jones, Aidan~N Gomez, {\L}ukasz Kaiser, and Illia Polosukhin.
\newblock Attention is all you need.
\newblock {\em Advances in neural information processing systems}, 30, 2017.

\bibitem{wang2022impact}
Lei Wang, Chuan-Feng Fu, Philip~Pong Wong, and Qi Zhang.
\newblock The impact of tourists’ perceptions of space-launch tourism: An extension of the theory of planned behavior approach.
\newblock {\em Journal of China Tourism Research}, 18(3):549--568, 2022.

\bibitem{zyma2022application}
Oleksandr Zyma, Lidiya Guryanova, Nataliia Gavkalova, Natalia Chernova, and Olga Nekrasova.
\newblock The application of machine learning methods in determining attractive development directions for tourism businesses.
\newblock {\em Intelektin{\.e} ekonomika. ISSN 1822-8011, 2022, T. 16, Nr. 1}, 2022.

\end{thebibliography}
}

\end{document}